\documentclass[letterpaper, 10 pt, conference]{ieeeconf}  

\IEEEoverridecommandlockouts                              
\usepackage{graphicx} 
\usepackage{epsfig} 
\usepackage{mathptmx} 
\usepackage{times} 
\usepackage{amsmath} 
\usepackage{amssymb}  
\usepackage{url} 
\usepackage[normalem]{ulem} 

\title{\LARGE \bf
Toward Efficient and Robust Biped Walking Optimization
}

\author{Nihar Talele and Katie Byl
\thanks{This work is funded by an NSF CAREER award (CMMI
1255018).}
\thanks{Nihar Talele and Katie Byl are with the Electrical Engineering Department at the University of California, Santa Barbara CA 93106
        {\tt\small nihar@umail.ucsb.edu},
        {\tt\small katiebyl@ucsb.edu}}%
}

\begin{document}

\maketitle
\thispagestyle{empty}
\pagestyle{empty}

\begin{abstract}
Practical bipedal robot locomotion needs to be both energy efficient and robust to variability and uncertainty.  In this paper, we build upon recent works in trajectory optimization for robot locomotion with two primary goals.  First, we wish to demonstrate the importance of (a)~considering and quantifying not only energy efficiency but also robustness of gaits, and (b)~optimization not only of nominal motion trajectories but also of robot design parameters and feedback control policies. As a second, complementary focus, we present results from optimization studies on a 5-link planar walking model, to provide preliminary data on particular trade-offs and general trends in improving efficiency versus robustness.  In addressing important, open challenges, we focus in particular on discussions of the effects of choices made (a)~in formulating what is always, necessarily only an approximate optimization, in choosing metrics for performance, and (b)~in structuring and tuning feedback control.

\end{abstract}

\section{Introduction}

Humanoid walking should be both energy efficient and robust to perturbations. This paper explores methods and trade-offs in achieving these two goals through local (gradient-based) optimization of a 5-link planar walking model. 

Both energy efficiency and robustness have long been goals for robot walking, and both mechanical design and control strategy play important roles in each objective. Below is a selective summary of relevant prior work. 

Toward improved mechanical design, biped robots built on passive dynamic principles drew significant attention over a decade ago~\cite{collins2005efficient}, but their success at reducing required energy has seemed to be coupled with fragile dynamics, yielding susceptibility to falls.  
Design of mechanical properties, i.e., lengths and mass distribution, clearly play an important role in enabling efficient legged locomotion, but they also arguably affect stability.

To improve controlled walking strategies, a range of work has focused on both trajectory optimization and control theory.
Trajectory optimization through direct collocation~\cite{von1992direct} is one promising approach.

In 1999, for example, Hardt et al. formulated the problem of minimizing energy of a planar 5-link biped, both with and without ankle torque, using DIRCOL software~\cite{hardt1999optimal} to solve a nonlinear optimization subject to contact constraints. 
Two years late, Paul et al. looked at simultaneous optimization of both mass distributions (robot design) and nominal motion trajectories, to be tracked via a simple proportional controller (with saturation limits), using simple neural networks to learn efficient locomotion~\cite{paul2001road}. 

In 2002, Westervelt and Grizzle highlighted the importance of optimizing walking motions while simultaneously guaranteeing asymptotic stability~\cite{westervelt2002design}, as opposed to a still-dominating paradigm of sequential design, first optimizing a nominal trajectory and subsequently adding feedback control in a more ad hoc way. As in~\cite{hardt1999optimal}, they also use DIRCOL, and they solve a sequential quadratic programming (SQP) problem to optimize the sum of $u^2$ across all four actuators.
Note that \cite{westervelt2002design} uses a hybrid zero dynamic (HZD) approach, which parameterizes joint trajectories on a monotonic, geometric variable. In a similar spirit, \cite{djoudi2005optimal} produce energy-optimal gaits for the 5-link walker using polynomial trajectories in which the gait is defined as $q(s)$, as a function of geometry rather than time, by solving for optimal polynomial coefficients. 

Various works have instead focused on optimizing robustness.
In ~\cite{dai2012optimizing}, Dai and Tedrake optimized a measure of robustness that quantifies variation from a nominal trajectory during rough terrain locomotion, for both the spring-loaded inverted pendulum (SLIP) and compass gait (CG) walker planar legged locomotion models. In \cite{saglam2014quantifying}, Saglam and Byl explored Pareto trade-offs between energy use and robustness, using a weighted metric that balances rewards for both low energy use and high mean-time to failure (aka mean first-passage time), using value iteration to optimize across a meshed approximation of the reachable state space for the system. 

Recent work by Hamed, Buss and Grizzle also focuses on robustness, tuning control parameters to ensure not only stable eigenvalues of the Jacobian of the period-one return map of limit-cycle walking but also reduced sensitivity of this Jacobian to parameter variation~\cite{hamed2016exponentially}. Here, they decouple the selection of a nominal periodic orbit from that of optimizing a parameterized controller, e.g., torques include both the necessary feedforward terms exactly compatible with the limit cycle of interest, along with some flavor of feedback law (e.g., perhaps but not necessarily HZD) that has no effect along the exact limit cycle trajectory.

Finally, a few other recent works emphasize applicability of legged locomotion optimization to an expanding range of problems.
Recent work by Ma, Hereid, Hubicki and Ames on the DURUS robot employs the HZD framework to optimize energy efficiency for stable 3D running~\cite{hereid20163d}.  Xi, Yesilevskiy and Remy employ direct collocation (DC) to optimize energetic cost for gaits without a prescribed sequence of foot contacts with the grounds~\cite{xi2016selecting}, and within our own group, we have used trajectory optimization to predict the theoretical cost of added mass in exoskeleton design~\cite{sovero2016initial} and to discover nonintuitive locomotion strategies for an underactuated, acrobot-based rolling system~\cite{bellegarda2017humanoids}.

In this paper, we focus on several related, open challenges in simultaneously optimizing for energetics and robustness. We highlight important choices made in differentiation and integration that improve speed and accuracy, since local optimization provides only approximate results. With an aim toward improving both energetics and robustness, we explore how variations in mass distribution affect metrics for each of these goals and observe a natural trade-off (between metrics) that results from tuning of feedback control. Our results demonstrate that choice of both mass distribution and feedback control structure have important, and apparently coupled, effects on both energy use and stability, providing evidence for the hypothesis that more comprehensive frameworks are needed for simultaneous optimization across system parameters and desired metrics.


The rest of this paper is organized as follows. Section~\ref{sec:model} describes the 5-link planar walking model we study, while Section~\ref{sec:trajopt} outlines our choices on optimization framework, feedback structure for subsequent control of trajectories and definition of cost metric. Results are given in Section~\ref{sec:results}, followed by discussion and conclusions in Section~\ref{sec:conclude}.






\section{Simulation Model}
\label{sec:model}

\begin{figure}[!ht]	
\centering
	\includegraphics[height = 1.8 in]{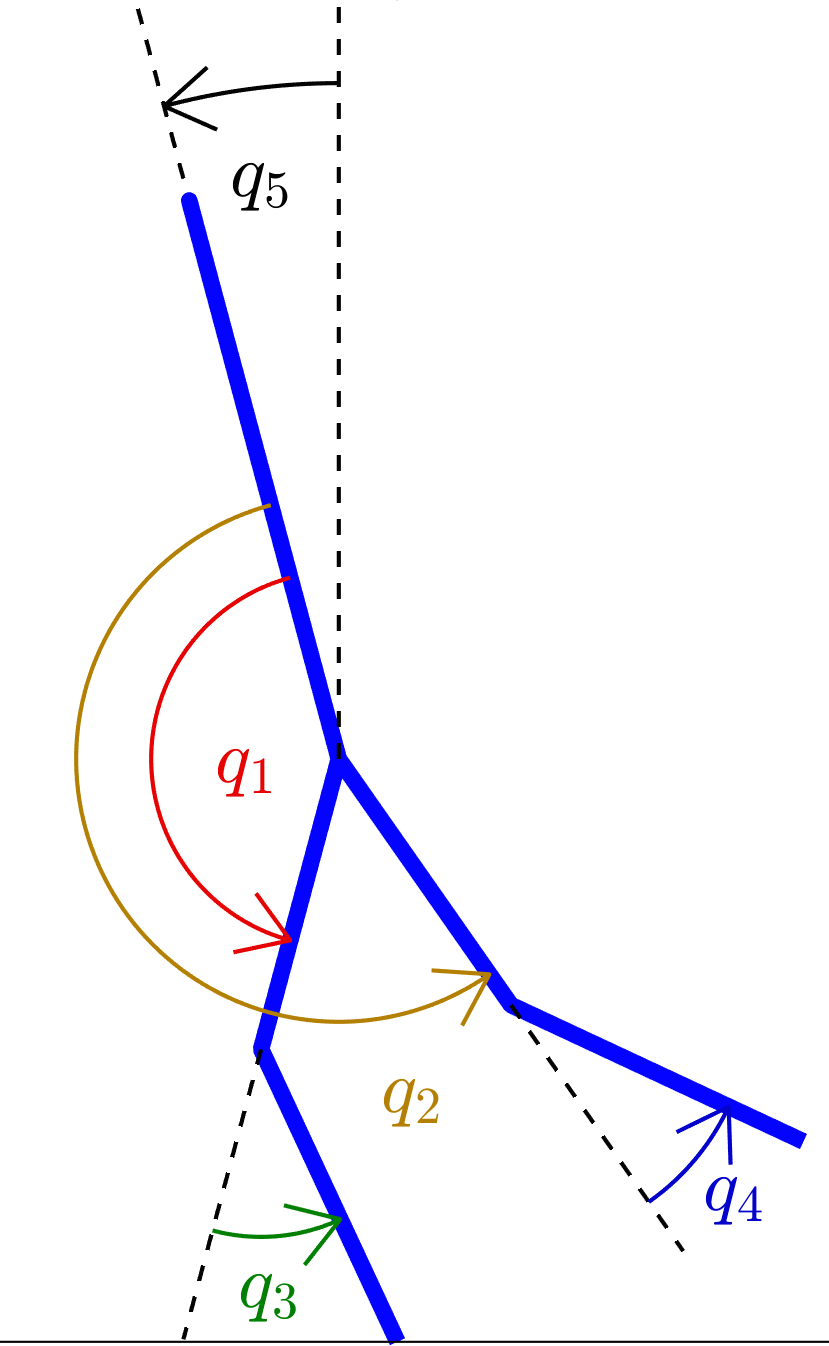} \hspace{.2in}
    \includegraphics[height = 1.8 in]{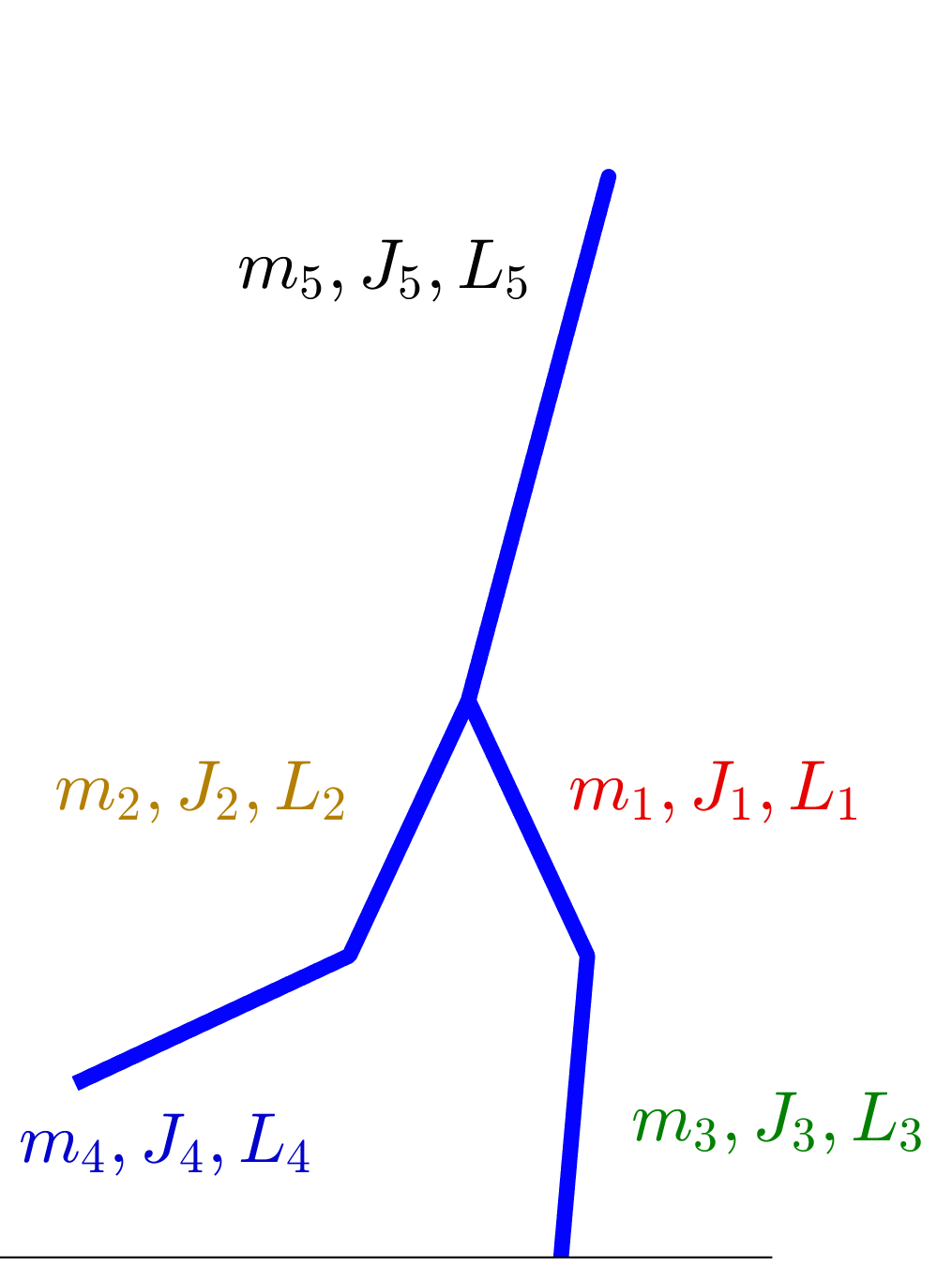}
\caption{5-link biped model. At left, $q_5$ is an absolute angle, measured with respect to vertical, while all other angles are relative. The lefthand image is drawn to clearly illustrate angles are positive in the counter-clockwise direction, throughout. At right, a typical pose, while walking to the right.} 
\label{fig:model_5link}
\end{figure}





We use a 5-link model, shown in Fig.~\ref{fig:model_5link}, for our simulations. The dynamics of this system are constrained to the sagittal plane only. We study several mass distributions, always enforcing that the total mass of the model is 70~(kg). The model has actuators at hips and knees, and the nonlinear dynamics can be written in the matrix form as 
\begin{equation}
\label{eq:dynamics}
D(q)\ddot{q} + C(q,\dot{q})\dot{q} + G(q) = Bu.
\end{equation}

This well-studied model~\cite{westervelt2002design} has 5 degrees of freedom, corresponding to 5 joint angles given by $q := [q_1~q_2~q_3~q_4~q_5]^\mathsf{T}$,  but due to the passive point-foot contact at the ground, the model still remains underactuated with $u \in \mathbb{R}^{4}$. We model the impact dynamics between the swing leg and the ground as instantaneous and inelastic~\cite{hurmuzlu1994rigid} to obtain joint velocities just after the impact. Since the impact model we use assumes inelastic collisions, some amount of energy is lost when the stance foot impacts the ground. 

\section{Trajectory Optimization}
\label{sec:trajopt}
\subsection{Framework for Direct Collocation}
We discretize the dynamics and use direct collocation (DC) to generate trajectories for the walking motion. Our approach is close to that suggested in~\cite{posa2013direct}, except that we use trapezoidal integration instead of backward Euler. Trapezoid rule integration can potentially result in lack of convergence. However, in our work, this had not been an issue, and results with trapezoid rule are significantly more accurate, when comparing the discretized (and thereby approximate) solutions inherent in this framework with subsequent high-resolution (1e-9) simulations of dynamics in Matlab.

The optimization problem is formulated as
\begin{align}
\mbox{find}\quad &q(t),~\dot{q}(t),~u(t)
\label{eq:optq}\\
\mbox{such that}\quad &D(q)\ddot{q} + C(q,\dot{q})\dot{q} + G(q) = Hu
\label{eq:opteom}\\ 
&\phi(q)=0
\label{eq:optphi}
\end{align}
where $ q \in \mathbb{R}^n$ is the vector of generalized coordinates, $ D(q) \in \mathbb{R}^{n\times n} $ is the mass inertia matrix, $C(q,\dot{q}) \in \mathbb{R}^{n\times n}$ represents the Coriolis forces, $G \in \mathbb{R}^n$ contains the gravitational forces and $H \in \mathbb{R}^{n\times {n-1}}$ is the input (torque) mapping. $\phi(q)$ is a vector of constraints. Constraints are imposed to make sure that the normal reaction at the point of contact with the ground is always positive. The optimization problem is set up such that at the end of the trajectory an impact at the ground happens. An additional constraint is added that the state of the model after the impact should match the initial condition in order to obtain a limit cycle behavior.


We implement this framework in Matlab making use of CasADi~\cite{Andersson2013b}, which lets us calculate gradients for optimization using algorithmic differentiation to machine precision. (CasADi uses \textit{C}omputer \textit{a}lgebra \textit{s}ystem syntax to perform \textit{Algebraic Differentiation}; thus the name.)
Using algebraic (and not numerical) differentiation greatly increases the stability and convergence properties of our optimization, while also reducing run time considerably.

Using CasADi to improve automated gradient calculation,
the nonlinear programming (NLP) optimization itself is solved using
IPOPT~\cite{biegler2009large}. This choice (vs use of SNOPT, Matlab's fmincon, etc.) is made based both on improved speed during our own in-lab testing experience and similar external benchmarking results~\cite{Mittleman2017Benchmarks}.

The DC framework evaluates Eqs.~\ref{eq:optq}-\ref{eq:optphi} only at discrete time intervals, $t_k$, resulting in an approximation of the desired optimization problem. We use $\Delta t = h = 0.01$~(s) and integrate using the standard trapezoid rule. Also, we assume $u(t)$ is held via a zero-order hold for each time step (as opposed to a first-order hold). Our integration scheme is then
\begin{align}
\label{eqtrapq}
\quad &q_{k+1} = q_k + \frac{h}{2}[\dot{q}_k + \dot{q}_{k+1}] \\
\label{eqtrapdq}
\quad &\dot{q}_{k+1} = \dot{q}_k + \frac{h}{2}[f(q_k,\dot{q}_k,u_{k}) + f(q_{k+1},\dot{q}_{k+1},u_{k})]
\end{align}
where $f(q,\dot{q},u) = D(q)\backslash(-C(q,\dot{q}) - G(q) + Bu)$.
We found that using a first order hold for input, i.e., replacing $u_k$ with $u_{k+1}$ at far right in (\ref{eqtrapdq}) above, leads to trajectories that are undesirably oscillatory and not smooth. This problem is easily rectified by adding some level of regularization. However, the zero order hold method still converges more rapidly. Also important to note is that regularization increases the optimal cost by a small amount.

\subsection{Trajectory Stabilization}
After we obtain the trajectories from the optimization framework, we simulate them in matlab using ode45 with an error tolerance of 1e-9. In order to stabilize these trajectories, we use partial feedback linearization (PFL). The total input to the system is $U = U_{ff} + U_{fb}$ where 
\begin{equation}
\label{eq:control_law}
U_{fb} = (SD^{-1}B)^{-1}(v + SD^{-1}(C\dot{q} + G)),
\end{equation} 
S is given by
\begin{equation}
\label{eq:mat_adjs}
S = \left[
\begin{tabular}{ccccc}
0 & 1 & 0 & 0 & 0\\
1 & 0 & 1 & 0 & 0\\
0 & 0 & 0 & 1 & 0\\
0 & 0 & 0 & 0 & 1\\ 
\end{tabular}
\right],
\end{equation}
and $u_{ff}$ contains the feedforward torques compatible with the nominal dynamics.

Given a passive contact of the stance leg with the ground, PFL allows us to directly set the accelerations of 4 out of 5 angles using $v$. We set $v = [\ddot{q}_{2des},\ddot{q}_{1des} + \ddot{q}_{3des},\ddot{q}_{4des},\ddot{q}_{5des}]^\mathsf{T}$ where $\ddot{q}_{des} = -K_p(q - q_{des}) - K_d(\dot{q} - \dot{q}_{des})$. We set $K_p = \omega_n^2$ and $K_d = 2\zeta\omega_n$. For all our simulations we set $\zeta = 1$ and test across a range of $\omega_n$ values. We get $U_{ff},~q_{des},~\dot{q}_{des}$ by interpolating the trajectories from the optimization framework.

\subsection{Cost of Transport (COT)}
We optimize the trajectories for cost of transport (COT) which is calculated as
\begin{equation}
\label{eq:COT}
COT =\frac{ \int_{0}^{t}\sum_{n=1}^{5}|P_n(t)|dt}{Mgd}
 \end{equation}
which we implement using trapezoid rule with zero order hold for input as
 
\begin{equation}
\label{eq:COT_disc}
COT =\frac{ \sum_{k=0}^{N-1}\sum_{n=1}^{5} \widetilde{P}_n(k) \Delta t}{Mgd}
 \end{equation}
where there are $N$ discrete time steps $t(k)$, $d$ is the stride length, and $M$ is the mass of the model. Rate of work (power) at joint $n$ during time step $k$ is approximated as
\begin{equation}
\label{eq:Pdef}
\widetilde{P}_n(k) = \frac{\widetilde{P}_{n,1}(k)+ \widetilde{P}_{n,2}(k)}{2} ~~~~~ n=\left\{1,2,...,5\right\},
\end{equation}
where $\widetilde{P}_{n,1}(k)$ is the regularized version of $P_{n,1}(k)=\tau_n(k)\omega_n(k)$ and $\widetilde{P}_{n,2}(k)$ is the regularized version of $P_{n,2}(k)=\tau_n(k)\omega_n(k+1)$. Also, $\widetilde{P}_{n,i}(k)$ is defined to penalize both the positive as well as the negative mechanical work. We smooth the cost function by using a regularization factor, so that
\begin{equation}
\label{eq:Ptilde}
\widetilde{P}_{n,i}=\sqrt[]{P_{n,i}^2 + \epsilon^2}~~~~~~~~~i=\left\{1,2\right\}
\end{equation}
as suggested in~\cite{srinivasan2006walk}. We use $\epsilon^2=0.01$, which works well for our optimization. Lower values of $\epsilon$ led to stability issues with the solver.

\section{Results}
\label{sec:results}

\begin{figure}[!ht]	
\centering
\includegraphics[height = 1.4 in]{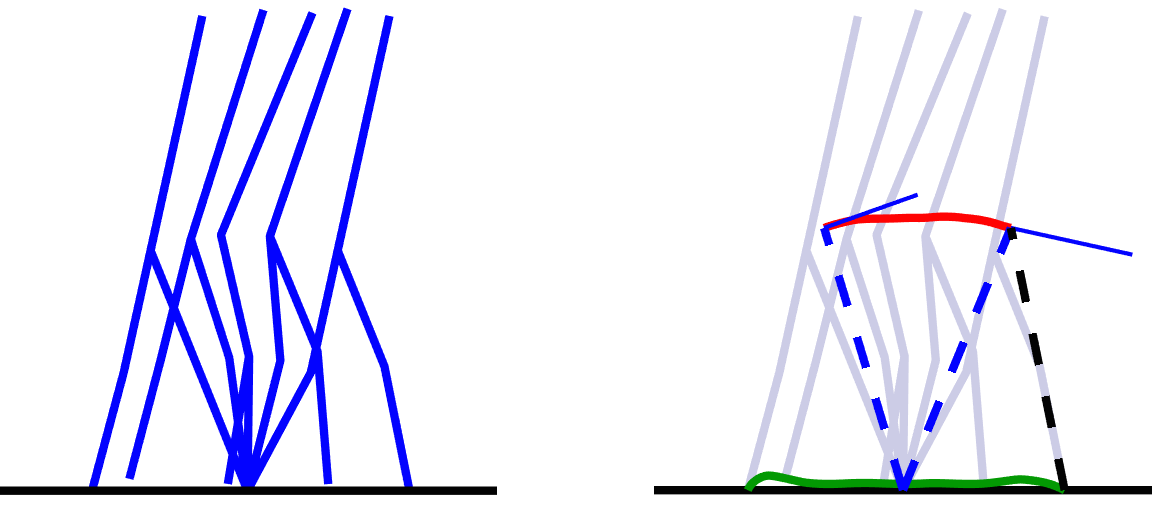}
\caption{Typical motion for the 5-link walker, using trajectories generated from the optimization framework. The figure on the left shows snapshots of motion. On the right, trajectories of the COM and the end of the swing foot are overlaid.} 
\label{fig:anim_5link}
\end{figure}

\begin{figure}[!ht]	
\centering
	\includegraphics[width = 3.0 in]{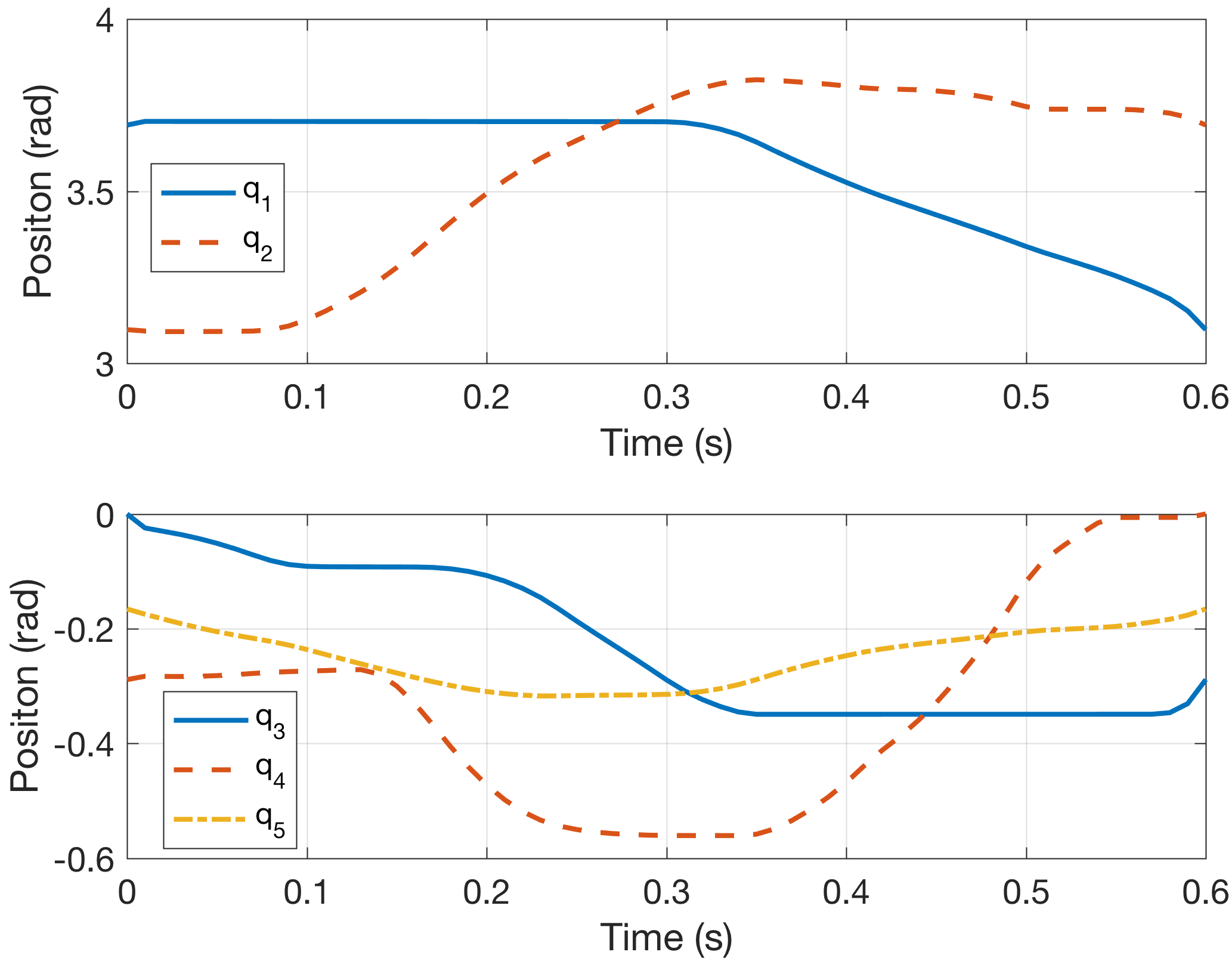}
\caption{Angle trajectories for the walking shown in Fig.~\ref{fig:anim_5link}. The top plot shows the positions of the swing and the stance thighs, while the bottom plot shows the positions of the stance knee, swing knee and torso.} 
\label{fig:traj_pos}
\end{figure}

\begin{figure}[!ht]	
\centering
\includegraphics[width = 3.0 in]{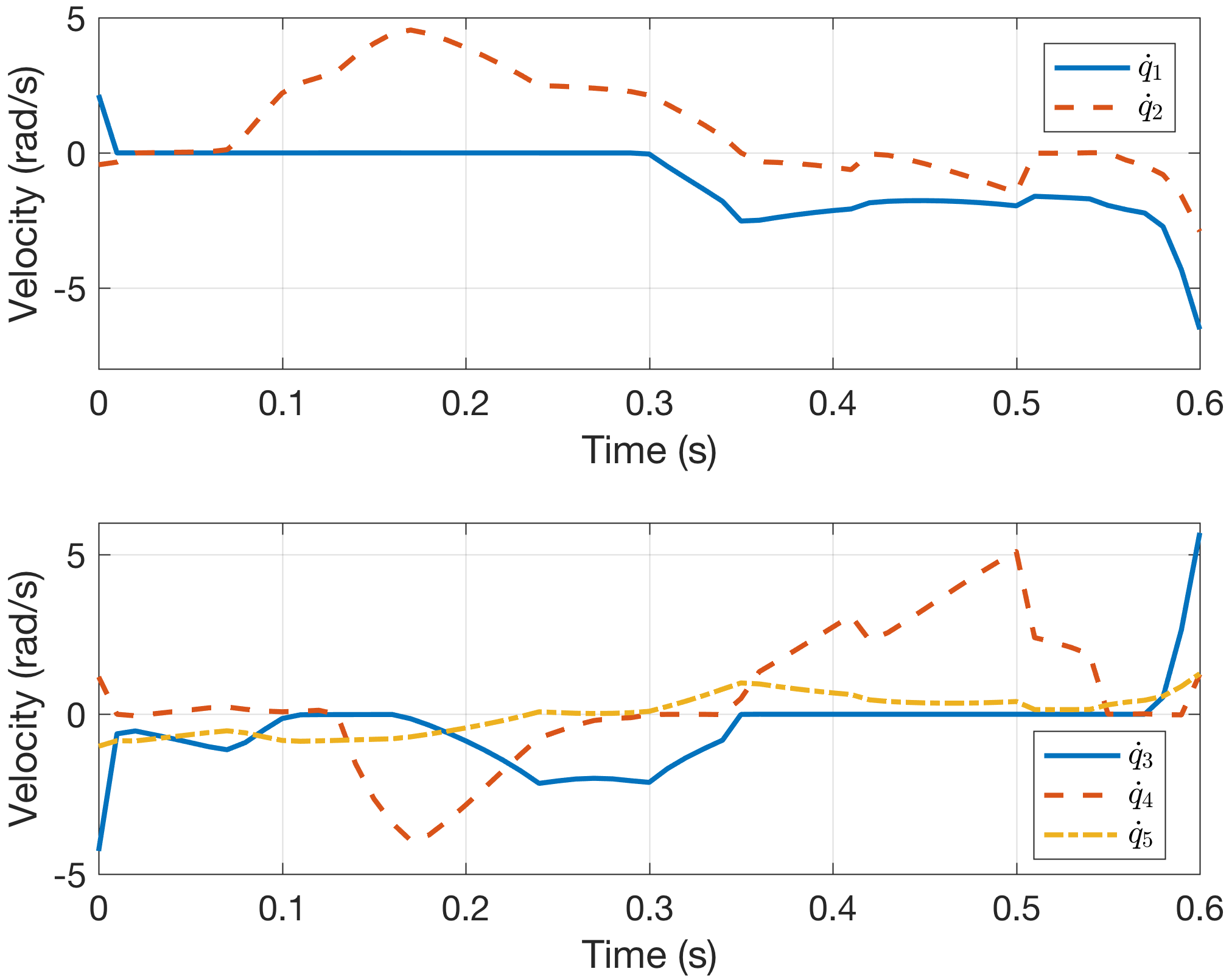}
\caption{Velocity trajectories for the walking motion shown in Fig.~\ref{fig:anim_5link}. At top are velocities of the swing and the stance thigh; bottom plot shows the velocities of the stance knee, swing knee and torso.} 
\label{fig:traj_vel}
\end{figure}

\begin{figure}[!ht]	
\centering
\includegraphics[width = 3.1 in]{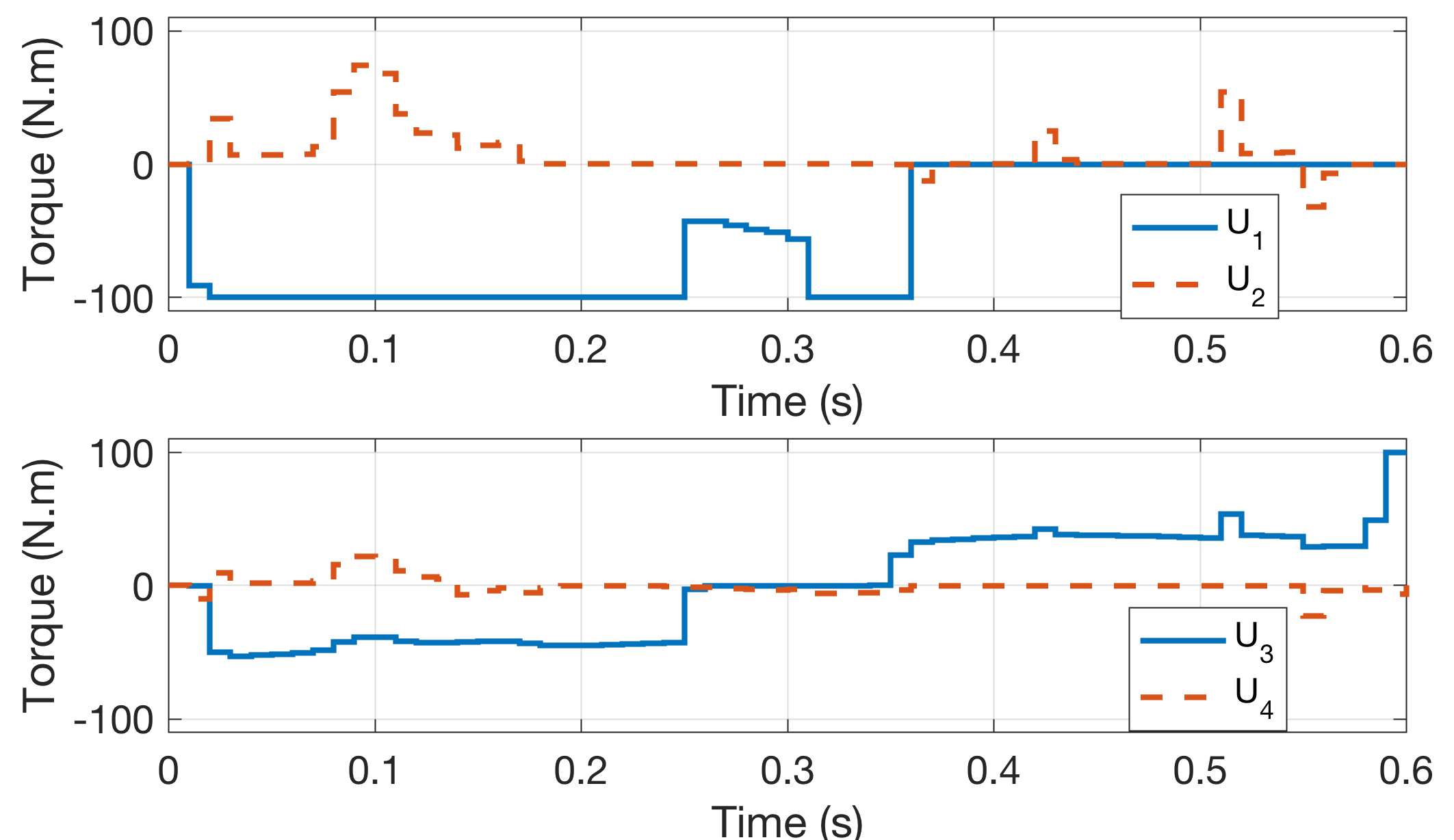}
    \caption{Torques for the walking motion in Fig.~\ref{fig:anim_5link}.}
\label{fig:traj_torque}
\end{figure}

Fig.~\ref{fig:anim_5link} shows a typical motion generated for the mass distribution $m_5 = 50~(kg)$, $m_1 = m_2 = 7~(kg)$, $m_3 = m_4 = 3~(kg)$, which corresponds roughly to a human mass distribution. Fig.~\ref{fig:traj_pos} and Fig.~\ref{fig:traj_vel} show the corresponding angle and angular velocity trajectories for that motion, and Fig.~\ref{fig:traj_torque} shows the joint torques. All the results we present here are generated for a walking motion of stride length~= 0.6~m over a time interval of 0.6~(s), resulting in a velocity of 1~(m/s).

A few details in these four figures are worth pointing out.  First, note that all trajectories are divided into two subplots for better resolution and clarity, since $q_1$ and $q_2$ remain close to $\pi$, while the other joints are near $0$. Upper plots correspond to upper leg segments; solid (blue) lines correspond to stance leg segments (femur and tibia). 

Several characteristics seen in this example are common among optimizations we performed across a range of mass distributions, as itemized below:
\begin{enumerate}
\item{The swing leg follows a very low trajectory, as depicted by the solid green line in Fig.~\ref{fig:anim_5link}. We enforce a minimum ground clearance of 2~cm, except at the first and last 5 points of the trajectory, and the swing leg tip overshoots near the ends and grazes this value mid-gait.
Without adequate ground clearance (e.g., 2~(cm)), resulting limit cycle gaits could not be stabilized.}
\item{The center of mass (COM) trajectory ``flattens out'' mid-stride, as opposed to following an arc, which is a feature also seen in human walking. This trajectory requires work in bending of the stance leg but results in less acceleration and deceleration vertically (against gravity), for lower accelerations overall of COM.}
\item{Also, the COM velocity at the end of gait is deflected slightly ``upward'', to reduce kinetic energy losses at impact. The velocity vector, depicted as a solid blue line in Fig.~\ref{fig:traj_vel}, is close to orthogonal with the dashed line in Fig.~\ref{fig:anim_5link} drawn from COM to stance leg tip at the start of the step (at $92.8^\circ$) and more obtuse at the end ($100.0^\circ$ with respect to current stance contact, and $66.0^\circ$ wrt upcoming stance leg, as a black dashed line).}
\item{Also, the velocity vector is longer (i.e., faster speed) at the end, showing kinetic energy has built up, to compensate for dissipation at impact.}
\end{enumerate}


The velocity (Fig.~\ref{fig:traj_vel}) and torque (Fig.~\ref{fig:traj_torque}) trajectories also show repeatable features that were not anticipated but are (retrospectively) intuitively in agreement with our cost function, as noted in the rest of the list of features, below. 

\begin{enumerate}
\setcounter{enumi}{4}
  \item{\textit{Magnitude} of velocities of the stance leg segments (solid blue) increase \textit{quite rapidly} at the very end of the step, in achieving the COM deflection upward (item 3).} 
  \end{enumerate}
The ``toe-off'' described above has a well-known benefit for energetics~\cite{kuo2002energetics}, but it also causes a problem:
\begin{enumerate}
\setcounter{enumi}{5}
  \item{We observe that rapid increases in velocity near the end of a step, required for toe-off, also have an important (and unfortunate) negative effect on stability.}
  \end{enumerate}
For example, when we attempted to use linear quadratic regulation (LQR) to provide feedback control, the resulting gait was not stable.  Specifically, we linearized about each ``knot point'' of the optimal solution, and then controlled motions using the nominal feedforward torque ($U_{ff}$) added to feedback of the form $U_{fb} = -K (X-X_{nom})$, where $K=K(t)$ and $X_{nom}=X_{nom}(t)$ were interpolated between their values at the discrete points of the optimal solution.  During continuous motion, the trajectories definitely converged toward the nominal trajectories as expected, but the effects through impacts were too destabilizing, resulting in falls after 3 to 6 steps.
  
This is a particularly interesting result, as it demonstrates evidence for a strong hypothesis that trajectories and feedback policies should be optimized as a concurrent problem, rather than planning ad hoc feedback subsequent to solving for a nominal trajectory. 

Finally, 
\begin{enumerate}
\setcounter{enumi}{6}
\item{we notice the values of $u_n$ and the corresponding relative angular velocity $\dot{q}_n$ show a complementary behavior: when one is significant in magnitude, the other is near zero.}
\end{enumerate}
This makes sense, given $P_n = u_n \dot{q}_n$, from Eq.~\ref{eq:Pdef}. In Figures~\ref{fig:traj_vel} and \ref{fig:traj_torque}, note in particular the solid blue lines, for the stance thigh (upper) and knee joints.  The relative thigh angle (between torso and stance leg) is nearly zero while torque is at its maximum magnitude (-100~N$\cdot$m), with the associated bobbing motion of the torso in Fig.~\ref{fig:anim_5link} during the first 0.25 seconds of the step.  For the stance knee, a period of negative velocity for $\dot{q}_3$ (i.e., knee bending) at mid-stance corresponds to a flat region in which $u_3 \approx 0$, i.e., bending almost passively during the gait.  Near the end of the step, push-off with the stance knee is concentrated in particular at the last time step, perhaps exploiting the approximate nature of the optimization problem somewhat.

A different choice of cost function would result in somewhat different solution characteristics.  We have also tested using a simple quadratic cost on control effort (i.e., cost = $u^2$), and we find the ``toe-off'' behavior, with a spike in torque and velocity of the stance knee at the end of the step, is a repeatable feature.  Overall, the trajectories are qualitatively more smooth for this cost function, however, and the overall COT is noticeably higher than when optimizing for COT specifically.

%

\subsection{Effect of system parameters on energy and stability}
\label{ss:sysparams}
We repeated the same optimization for a total of five sets of model parameters. The link lengths, shown in Table~\ref{tab:length_params}, remained the same in all trials. Table~\ref{tab:mass_params} shows the masses for each set. Each link is modeled as a simple rod, with COM at mid-length.

\begin{table}[ht]
\begin{center}
\begin{tabular}{|c|c|c|}
\hline
Segment & $Length(m)$ & $COM(m)$ \\
\hline
$L_1 = L_2$ & 0.4 & 0.2 \\
\hline
$L_3 = L_4$ & 0.43 & 0.215 \\
\hline
$L_5$& 0.77 & 0.385 \\
\hline
\end{tabular}
\end{center}
\caption{Length and COM parameters used in simulation experiments}
\label{tab:length_params}
\end{table}

\begin{table}[ht]
\begin{center}
\begin{tabular}{|c|c|c|c|c|}
\hline
Set & $m_1 = m_2(kg)$ & $m_3 = m_4(kg)$ & $m_5(kg)$ & $COT_{opt}$\\
\hline
1 & 7 & 7 & 42 & 0.0992\\
\hline
2 & 5 & 7 & 46 & 0.0996\\
\hline
3 & 7 & 5 & 46 & 0.0861\\
\hline
4 & 5 & 5 & 50 & 0.0853\\
\hline
5 & 7 & 3 & 50 & 0.0705\\
\hline
\end{tabular}
\end{center}
\caption{Mass parameters used in simulation experiments}
\label{tab:mass_params}
\end{table}

\begin{figure}[!ht]	
\centering
	\includegraphics[width = 3.1 in]{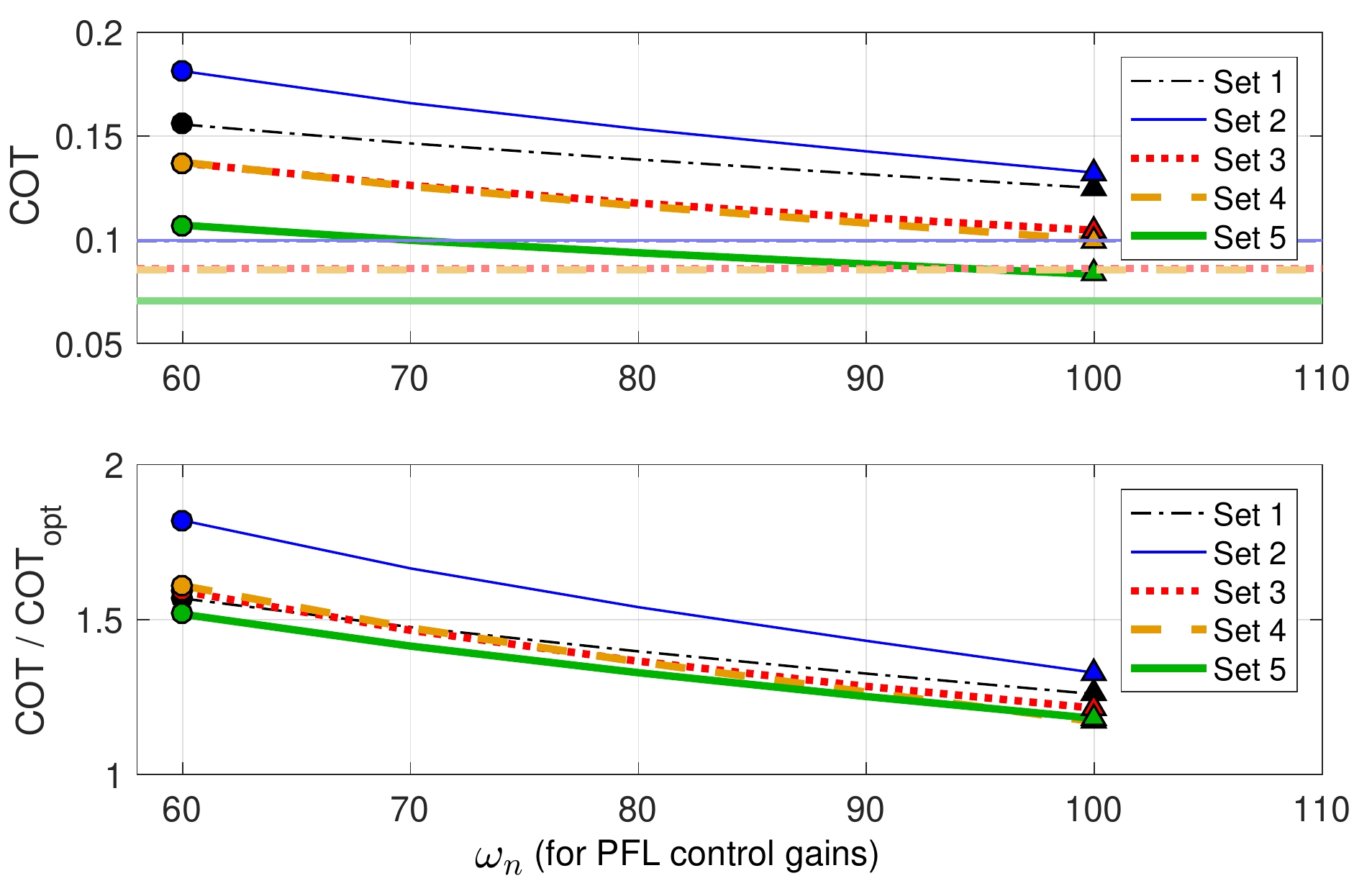}
\caption{Cost of Transport (COT) as a function of $\omega_{n}$.} 
\label{fig:COTvsomega}
\end{figure}

Fig.~\ref{fig:COTvsomega} shows how changing $\omega_n$ affects COT. First, recall that discretized trajectory optimization solutions of true dynamics are always approximate (\cite{von1992direct,hardt1999optimal,hamed2016exponentially,xi2016selecting,bellegarda2017humanoids,posa2013direct,srinivasan2006walk}, etc.), and also that $\omega_n$ is analogous to a ``stiffness'' (or proportional controller, ``$K_p$''), within the PFL framework. Intuitively, increasing control gains would increase control effort, in following an arbitrary reference trajectory.

However, our reference trajectories are far from arbitrary: they are precalculated to achieve a locally optimal (i.e., minimal) cost of transport~-- within some approximation errors. Therefore, increasing $w_n$ monotonically decreases resulting COT in our higher-accuracy Matlab simulations.


Figure~\ref{fig:COTvsomega} both illustrates how COT converges exponentially downward as $\omega_n$ increases and also how different parameter sets result in a range of different errors, in comparing optimization results to more accurate simulations. 
For example, the lowest COT is for Set 5 from Table~\ref{tab:mass_params}, which is also closest to a typical human mass distribution. The lower subplot of Fig.~\ref{fig:COTvsomega} shows the ratio of ``actual'' COT (from simulation) to the value output from optimization, also as a function of $\omega_n$, where Set 5 also shows the lowest approximation error.

\begin{figure}[!ht]	
\centering
	\includegraphics[width = 2.8 in]{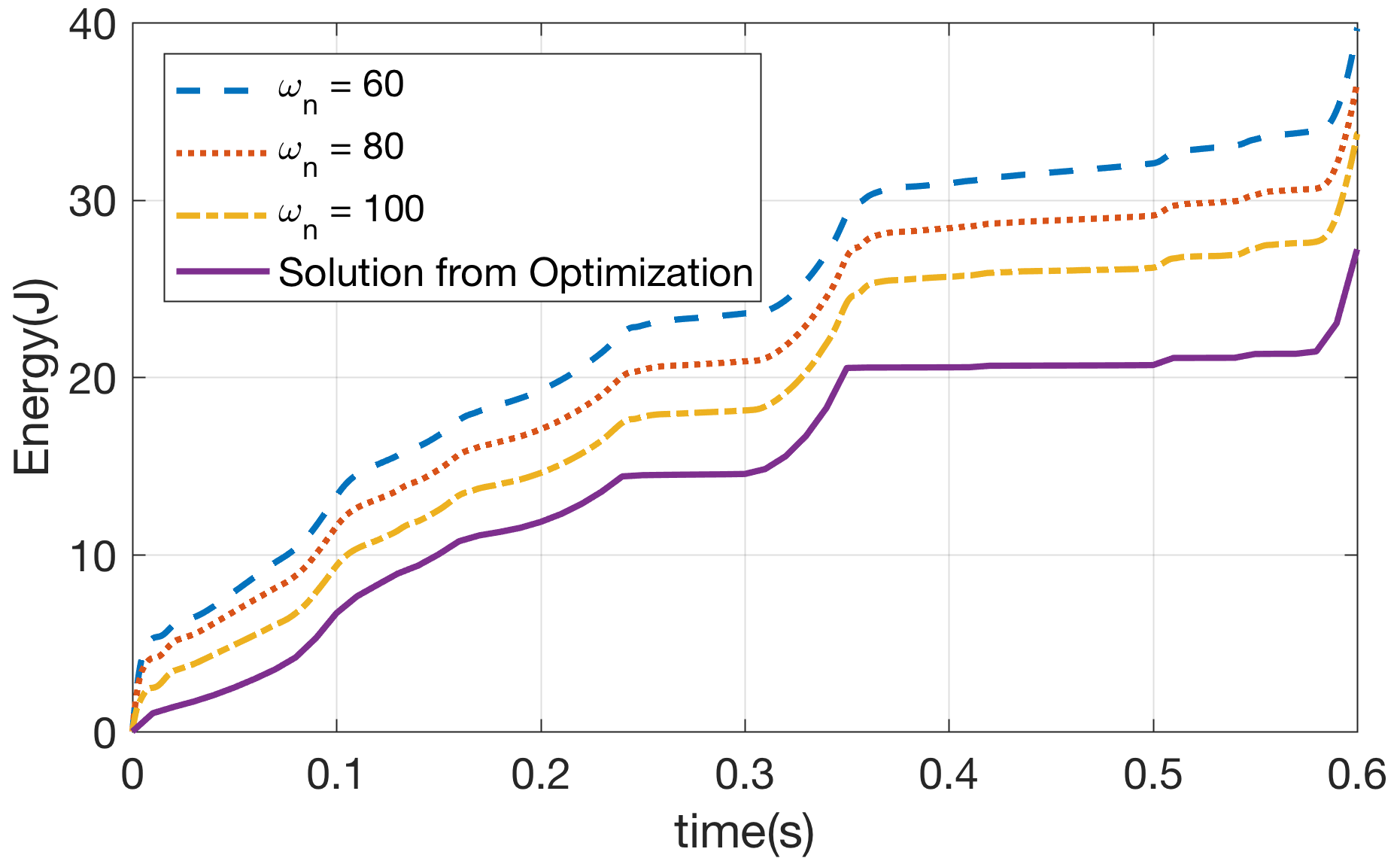}
\caption{Variation of energy as $\omega_n$ changes.} 
\label{fig:energyVSomega}
\end{figure}

Figure~\ref{fig:energyVSomega} shows the evolution of energy over time for different values of $\omega_n$. The steep rise at the start is due to high peak torques as the controller tries to pull the actual trajectory back to the optimal trajectory.

\subsection{Energy vs Stability Tradeoff}
\begin{figure}[!ht]	
\centering
\includegraphics[width = 3.3 in]{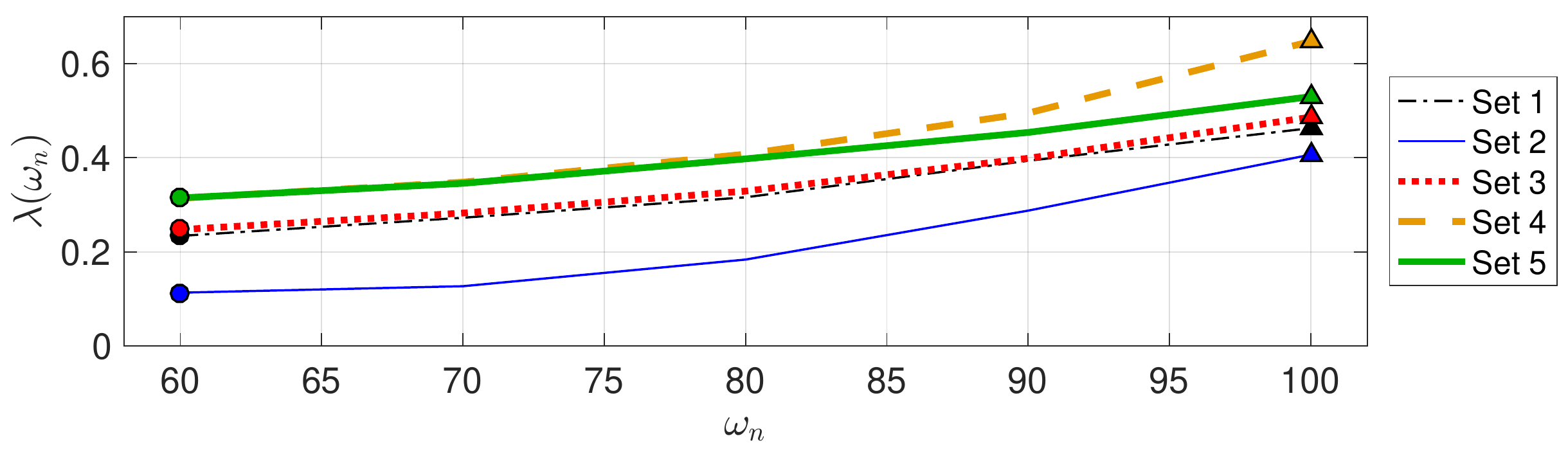}
\caption{Rate of convergence, $\lambda$, as a function of $\omega_{n}$, for push recovery.} 
\label{fig:eigVSomega_ave}
\end{figure}

Fig.~\ref{fig:eigVSomega_ave} shows the change in rate convergence (i.e., largest/slowest discrete-time eigenvalue) as a function of $\omega_n$ when recovering from push perturbations. Three impulsive push tests were simulated for each set of parameters by applying an impulse of 10~Ns (e.g., effectively a pulse of 1000N for 0.01s) at the hip, stance knee, or torso, respectively. The velocities post impact were calculated by using
\begin{equation}
\label{eqn:push}
D(q)\dot{q}^+ - D(q)\dot{q}^- = E \cdot F_{ext} \Delta t
\end{equation}
where $F_{ext} \Delta t = 10$~(Ns) is the applied pulse and $E = \frac{\partial{p}(q)}{\partial{q}}$ where $p(q)$ is the point at which force is applied. 

As with Figure~\ref{fig:COTvsomega}, the trend seen in Figure~\ref{fig:eigVSomega_ave} as a function of $\omega_n$ is not immediately intuitive. Usually, it would be expected that higher values of $\omega_n$ should lead to faster rates of convergence as higher gains should pull back the trajectory to the optimal reference trajectory faster. 

\begin{figure}[!ht]	
\centering
	\includegraphics[width = 3 in]{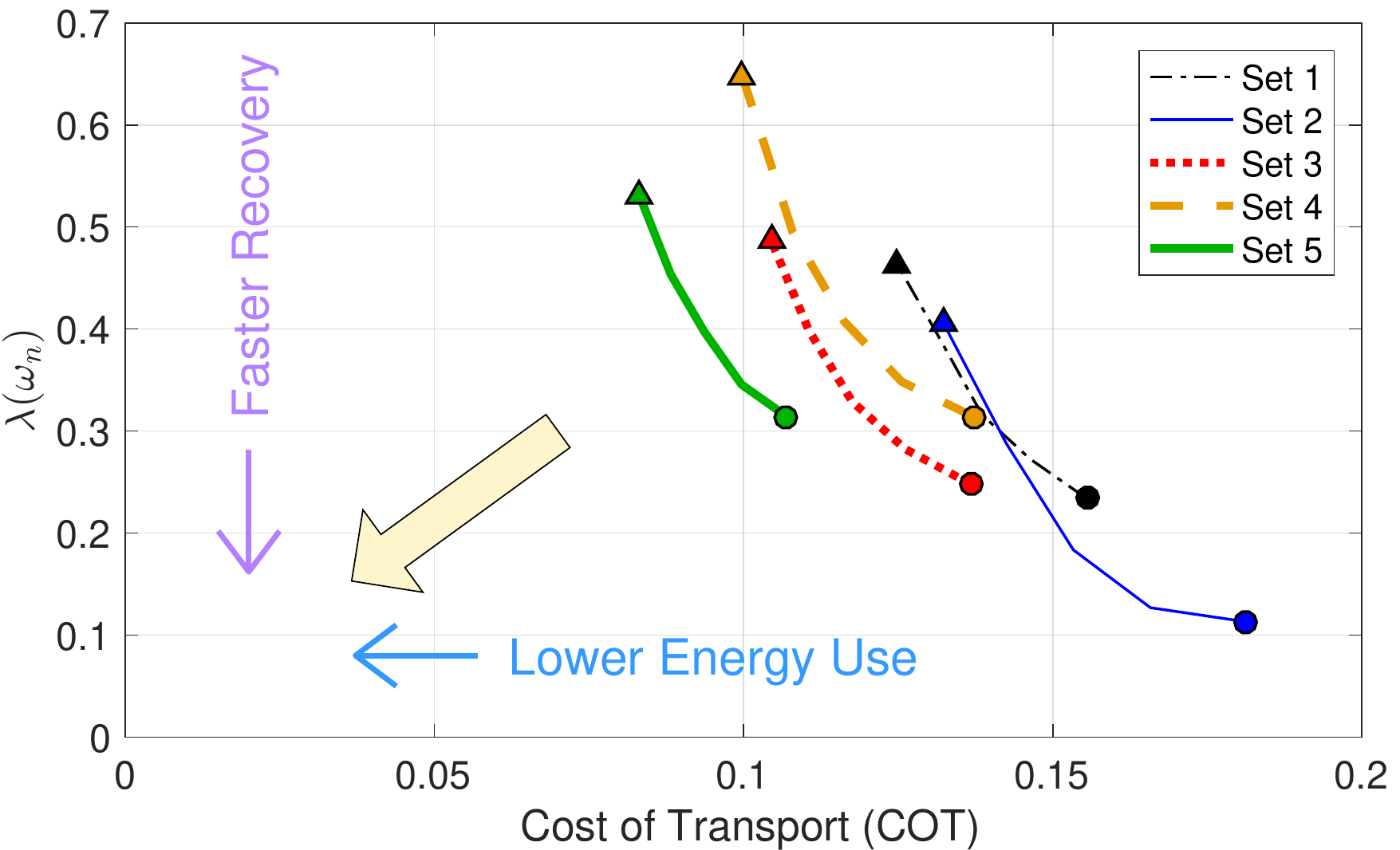}
\caption{Cost of Transport vs rate of convergence $\lambda(\omega_n)$, for push recovery. Each line sweeps across results from $\omega_n=60$ to $\omega_n=100$.} 
\label{fig:cotVSeig_ave}
\end{figure}

Combining data from Figures~\ref{fig:COTvsomega} and \ref{fig:eigVSomega_ave} to investigate any potential relationship between stability and energy further, Fig.~\ref{fig:cotVSeig_ave} shows a plot of COT vs rate of convergence for the parameter sets given in Table~\ref{tab:mass_params}.  A clear trade-off between energy efficiency and rate of convergence is evident here.  

\section{Discussion and Conclusions}
\label{sec:conclude}

Our simulation results demonstrate that although optimizing for energy alone and then stabilizing the trajectories can work, there is a need for a more cohesive framework that takes into account both energy as well as stability of the system for optimization simultaneously.  Furthermore, because different physical robot design parameters have important effects on stability and energy properties, it would be prudent to include these parameters as open variables, via appropriate optimization frameworks.

Toward better understanding of the effects of system parameters on the energy and stability of biped systems, we presented simulation data for 5 different sets of parameters. For the mass distribution sets we chose, the one which is most similar to human parameters (Set 5) does in fact have the lowest energy consumption. However, the set with the fastest rate of convergence is set 2. These phenomena may in part be a result of other factors, such as choice of feedback control structure and certainly warrant further study. 

\begin{figure}[!th]	
\centering
	\includegraphics[width = 2.3 in]{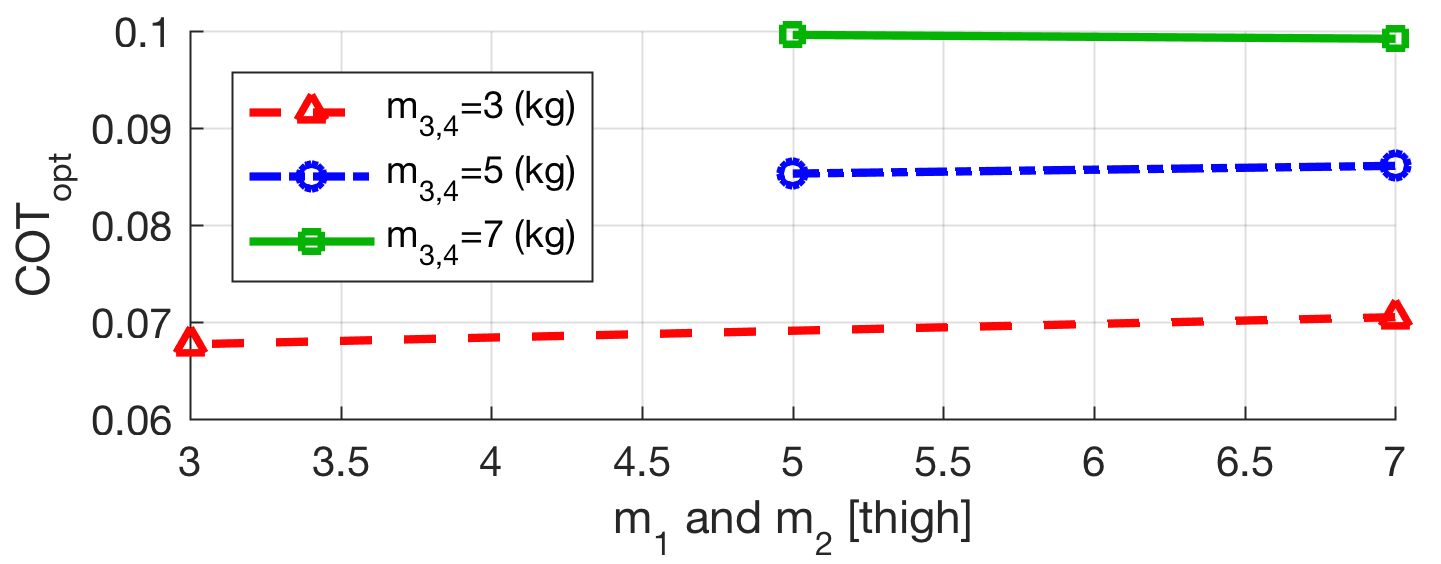}
    \includegraphics[width = 2.3 in]{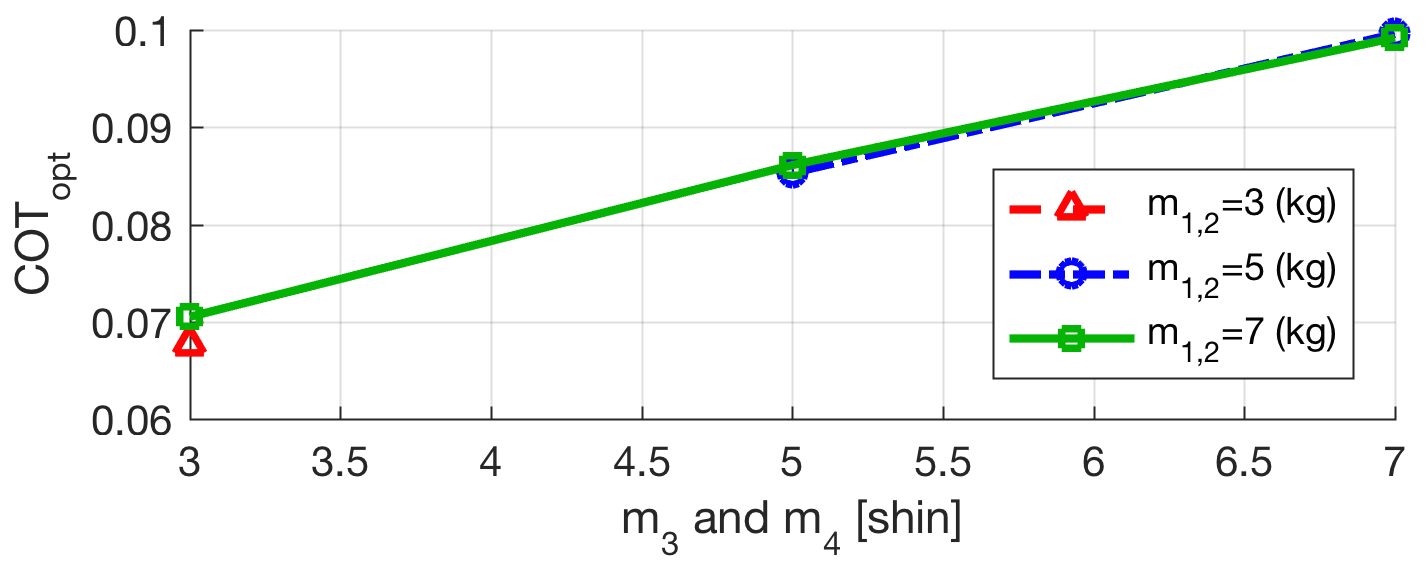}
\caption{COT variations as upper or lower leg mass varies.} 
\label{fig:205}
\end{figure}

Also, these and other simulations we have performed show that while increasing the mass of the lower leg (with constant upper-leg mass and total mass) results in a fairly linear increase in COT, as shown in the lower subplot of Figure~\ref{fig:205}, energy use remains close to flat as mass of the upper leg increases (while holding lower-leg and total mass constant). Corresponding trends relating mass distribution to stability are less apparent and deserve further investigation.

To explore trade-offs, we presented COT vs rate of convergence in Figure~\ref{fig:cotVSeig_ave}, which illustrates a Pareto frontier, formed essentially by sets 2, 3, and 5, while sets 1 and 4 provide poor trade-off characteristics by comparison. We hypothesize that a more comprehensive framework that includes energy, robustness, and physical parameters in a single optimization can improve the limits of such performance trade-offs, and developing such a framework is a goal for our future work.

\section*{Acknowledgment}
The authors thank Prof. Lars Struen Imsland of the Department of Engineering Cybernetics at NTNU for very helpful discussions on various optimization framework choices, toward improving both speed and accuracy.


\bibliographystyle{IEEEtran}
\bibliography{humanoids2017}

\end{document}